\documentclass[10pt,twocolumn,letterpaper]{article}

\usepackage[pagenumbers]{cvpr} 

\usepackage{graphicx}
\usepackage{amsmath}
\usepackage{amssymb}
\usepackage{booktabs}
\usepackage{gensymb}
\usepackage{multirow}
\usepackage{algorithmic}
\usepackage{algorithm}
\usepackage{scalefnt}

\usepackage{pgfplots}
\pgfplotsset{width=10cm,compat=1.9}
\definecolor{Seaborn1}{HTML}{4C72B0}
\definecolor{Seaborn2}{HTML}{DD8452}
\definecolor{Seaborn3}{HTML}{55A868}
\definecolor{Seaborn4}{HTML}{C44E52}
\definecolor{Seaborn5}{HTML}{8172B3}
\definecolor{Seaborn6}{HTML}{937860}
\definecolor{Seaborn7}{HTML}{DA8BC3}
\definecolor{Seaborn8}{HTML}{8C8C8C}
\definecolor{Seaborn9}{HTML}{CCB974}
\definecolor{Seaborn10}{HTML}{64B5CD}

\pgfplotsset{
compat=1.9,
legend image code/.code={
\draw[mark repeat=4,mark phase=1]
plot coordinates {
(0cm,0.02cm)
(0.1cm,0.02cm)        
(0.2cm,0.02cm)
(0.3cm,0.02cm)       
(0.4cm,0.02cm)      
};%
}
}

\newcommand{\V}[1]{\mathbf{#1}}
\newcommand{\diag}{\textrm{diag}}

\newcommand\nocell[1]{\multicolumn{#1}{c|}{}}

\usepackage[pagebackref,breaklinks,colorlinks]{hyperref}

\usepackage[capitalize]{cleveref}
\crefname{section}{Sec.}{Secs.}
\Crefname{section}{Section}{Sections}
\Crefname{table}{Table}{Tables}
\crefname{table}{Tab.}{Tabs.}

\def\cvprPaperID{12179} 
\def\confName{CVPR}
\def\confYear{2024}

\begin{document}

\title{Robust Self-calibration of Focal Lengths from the Fundamental Matrix}

\author{Viktor Kocur$^{1,2}$
\quad
Daniel Kyselica$^{1}$
\quad
Zuzana K\'ukelov\'a$^{2}$\\
$^{1}$ Faculty of Mathematics, Physics and Informatics, Comenius University in Bratislava \\
$^{2}$ Visual Recognition Group,
Faculty of Electrical Engineering, Czech Technical University in Prague \\
{\tt\small \{viktor.kocur, daniel.kyselica\}@fmph.uniba.sk \, kukelzuz@fel.cvut.cz}}

\maketitle

\begin{abstract}
The problem of self-calibration of two cameras from a given fundamental matrix is one of the basic problems in geometric computer vision. Under the assumption of known principal points and square pixels, the well-known Bougnoux formula offers a means to compute the two unknown focal lengths. However, in many practical situations, the formula yields inaccurate results due to commonly occurring singularities. Moreover, the estimates are sensitive to noise in the computed fundamental matrix and to the assumed positions of the principal points.
 
 In this paper, we therefore propose an efficient and robust iterative method to estimate the focal lengths along with the principal points of the cameras given a fundamental matrix and priors for the estimated camera parameters. In addition, we study a computationally efficient check of models generated within RANSAC that improves the accuracy of the estimated models while reducing the total computational time. Extensive experiments on real and synthetic data show that our iterative method brings significant improvements in terms of the accuracy of the estimated focal lengths over the Bougnoux formula and other state-of-the-art methods, even when relying on inaccurate priors.

\end{abstract}

\section{Introduction}
\label{sec:intro}

Camera calibration, \ie, the estimation of camera intrinsic parameters,  is a fundamental problem in computer vision with many applications. The precision of estimated intrinsic parameters, such as focal length and principal point, significantly affects the precision of tasks in structure-from-motion (SfM)~\cite{snavely2008modeling},visual localization~\cite{Sattler-PAMI-2017}, 3D object detection~\cite{xu_3dobj}, augmented reality~\cite{Castle08ISWC}, and other applications~\cite{sochor_vehiclespeed}. 
 
 Classical calibration methods~\cite{zhang_calibration} use known calibration patterns, \eg checkerboards, or additional knowledge of the observed scene to estimate the camera intrinsics. As such, they are often impractical.
 On the other hand, self-calibration
 methods do not require any additional knowledge of the scene geometry and 
rely on automatic detection of image features in input cameras. Thus, they are preferable in many applications.

 In this paper, we study the problem of self-calibration of two cameras, with potentially different intrinsic parameters that capture the same scene. Although this is a well-studied problem, with first solutions dated back to the nineties, to the best of our knowledge, all its existing solutions suffer from some instabilities and robustness problems.

The geometry of two uncalibrated cameras is captured by the fundamental matrix~\cite{fundamental_matrix}. In 1998 Bougnoux~\cite{bougnoux_original} showed
that assuming square pixels and known principal points of both cameras, it is possible to obtain the focal lengths of the two cameras from the fundamental matrix using a closed-form formula.
This provides a convenient way to determine the focal length, since the principal point can usually be assumed to lie in the image center. However, the precision of the estimated focal lengths using the Bougnoux formula~\cite{bougnoux_original} and other similar methods~\cite{kanatani_closed_form, melanitis, fetzer} is often marred by inaccuracies, sometimes even resulting in physically implausible imaginary focal lengths. This stems from singularities and susceptibility of these methods to noise in fundamental matrices and assumed positions of principal points. Moreover, the singularities occur in very common camera configurations when the principal axes of the cameras are coplanar (\ie they intersect), which is common when images of a single object of interest are taken from different views.

An alternative approach~\cite{hartley_iterative} relies on iterative optimization of a multiterm loss, allowing one to estimate both the focal lengths of the cameras as well as their principal points. This approach addresses some of the drawbacks of the aforementioned methods but still relies on the Bougnoux formula~\cite{bougnoux_original} within the iterative process. It is thus susceptible to inaccuracies, especially in the 
aforementioned degenerate camera configurations.

Similarly to \cite{hartley_iterative}, we formulate the problem of estimating focal lengths and principal points from a given fundamental matrix as a constrained optimization problem using priors. Constraints ensure that the solution satisfies the Kruppa equations~\cite{lourakis_calib}. 
To solve the constrained optimization problem, we search for all stationary points of its Lagrangian. 
This results in a complex system of polynomial equations that we efficiently solve in an iterative way. 
Each iteration in our formulation involves solving a system of two polynomial equations of degree four in two variables, which is efficiently solved using the Gr\"{o}bner basis method~\cite{larsson_efficient}.

Existing approaches, as well as our method, are based on fundamental matrices that can be estimated using different variants of RANSAC~\cite{ransac, magsac++, lo-ransac, degensac, vsac, prosac}. Some of the matrices produced by the 7-point algorithm~\cite{HZ-2003} within RANSAC lead to imaginary focal lengths when decomposed using the Bougnoux formula~\cite{bougnoux_original}. We observed that such matrices seldom lead to the final model. Based on this insight, we propose to perform a degeneracy check for such matrices and reject them, thus saving computational time on their further processing, such as scoring and local optimization. 

In summary, the contributions of this paper are:
\begin{itemize}
\vspace{-1.2ex}
    \item A novel iterative method for focal length and principal point estimation from fundamental matrices. Evaluation on synthetic and large-scale real-world datasets~\cite{phototourism_benchmark, aachen_dn_extended} shows that our method results in superior accuracy of estimated focal lengths as well as camera poses.
    \vspace{-1.2ex}
    \item A simple and computationally efficient degeneracy check for fundamental matrices leading to imaginary focal lengths within RANSAC. Performance on real-world datasets~\cite{phototourism_benchmark, aachen_dn_extended} shows that performing this check and rejecting such models within various RANSAC variants \cite{magsac++, lo-ransac, degensac, prosac} and implementations \cite{poselib, opencv, vsac} leads to faster computation and more accurate 
    pose and focal length estimates.
\end{itemize}

The code for the methods and experiments is available online.\footnote{\url{https://github.com/kocurvik/robust\_self\_calibration}}

\section{Related Work}
The problem of self-calibration of a camera from a given fundamental matrix  is a well-studied problem, with several solutions that can be divided into two main groups, the direct methods and the iterative methods.
\subsection{Direct Methods}

Bougnoux~\cite{bougnoux_original} showed that under the assumption of square pixels and known positions of the principal points, it is possible to calculate the focal lengths of both cameras from the fundamental matrix in  closed form. Kanatani and Matsunuga~\cite{kanatani_closed_form}  derived a closed form solution directly from the elements of the fundamental matrix, avoiding the intermediate epipole computation required by the Bougnoux formula. Melanitis and Maragos~\cite{melanitis} formulated a linear system, the solution of which provides the focal lengths.

These approaches are all algebraically equivalent and thus suffer from the same generic singularity, which occurs whenever the principal axes of the two cameras are coplanar, \ie they are intersecting. This is a very common scenario in practice, as human photographers tend to place objects of interest in the centers of images.
The singularity significantly affects the accuracy of the estimated focal lengths even when the axes are close to coplanar.

Similar formulas have been derived for the case when the focal lengths of cameras are known to be equal~\cite{sturm_single, kanatani_closed_form, brooks_single}. In case of equal focal lengths, it is possible to avoid estimating the full 7-DoF fundamental matrix and instead employ a 6-point minimal solver~\cite{stewenius_single_6pt} within RANSAC to find the relative poses of cameras along with the focal length. These methods also suffer from generic singularities when the principal axes are parallel, or they intersect, and the camera centers are equidistant from the intersection point~\cite{sturm_degen}.

An interesting case is when only one of the focal lengths is unknown. The focal length can be estimated in a closed form solution from the fundamental matrix~\cite{urbanek_onefocal} or using a 6-point minimal solver~\cite{bujnak_onefocal}. A degenerate case for these methods only occurs when the center of the calibrated camera lies on the principal axis of the uncalibrated camera, a situation rarely occurring in practice. 

Since focal lengths appear in squared form in the Kruppa equations~\cite{lourakis_calib}, all the previously mentioned direct methods provide only the squares of focal lengths. This may result in physically implausible imaginary focal lengths due to noise. Depending on the level of noise and camera configuration, imaginary focal lengths may occur at a significant rate (over 20\% of samples)~\cite{kanatani_stabilizing}. Imaginary focal lengths are less likely to occur when self-calibration is performed using three views~\cite{kanatani_threeview}.

\subsection{Iterative Methods}

Hartley and Silpa-Anan~\cite{hartley_iterative} pointed out that errors in the assumed positions of principal points may significantly affect the focal lengths estimated by the Bougnoux formula. To remedy this issue, they proposed an iterative method based on Levenberg-Marquardt optimization~\cite{lm}. During the optimization the fundamental matrix, along with the positions of principal points in both images, are optimized to minimize a multiterm loss. The loss consists of the Sampson error on the correspondences and the distance of the principal points and squared focal lengths from predetermined priors. To prevent imaginary focal lengths, the loss also contains a penalization term which dramatically increases the loss whenever the square focal lengths are below a given threshold. During the optimization procedure, the focal lengths are estimated from the fundamental matrix using the Bougnoux formula. This approach overcomes many of the issues of using the formula alone but may still suffer in the vicinity of degenerate camera configurations, as it still relies on the formula to estimate the focal lengths. Additionally, the procedure is computationally expensive as it requires computation of the Sampson error for all correspondences in each iteration.

Iterative methods can also be used to estimate the camera intrinsics using two or more views~\cite{gherardi_calib, pollefeys_calib, lourakis_calib, fetzer}. When considering only two views with two unknown focal lengths, these methods should converge to the Bougnoux formula. However, these methods may result in different outcomes in the vicinity of degenerate configurations. For example, the method by Fetzer \etal~\cite{fetzer} optimizes for an energy functional based on the Kruppa equations~\cite{lourakis_calib}. In degenerate configurations, the functional does not have a single minimum, yet the optimization procedure will be terminated at some point, providing an output different from the one provided by the Bougnoux formula.

Next, we describe the proposed iterative method that aims to avoid the aforementioned problems of existing approaches for self-calibration from the fundamental matrix.  

\section{Robust focal length estimation}
In this paper, similar to Hartley~\cite{hartley_iterative}, we formulate the problem of estimating focal lengths from a given fundamental matrix $\V F$ as an optimization problem using priors.

Let $\V F = \V U \V D \V V^{\top}$ be the SVD of the fundamental matrix $\V {F}$, where $\V{U} = [\V u_1, \V u_2, \V u_3]$  and $\V{V} = [\V v_1, \V v_2, \V v_3]$ are orthonormal matrices and $\V D = \diag (\sigma_1, \sigma_2, 0)$ is a diagonal matrix with two non-zero singular values $\sigma_1$  and $\sigma_2$ of $\V F$.
Let $\omega^{*}_i = \V K_i \V K_i^{\top}$ be the dual image of the absolute conic for the $i^{th}$ camera, $i = 1,2$, with $\V {K}_i$ being the $3 \times 3$ calibration matrix of the form 
\begin{equation}
\V K_i=
  \begin{bmatrix}
    f_i & 0 & u_i \\
    0 & f_i & v_i \\
    0 & 0 & 1 \\
  \end{bmatrix},
  \label{eq:K} 
\end{equation}
with the focal length $f_i$ and principal point~$\V{c}_i = [u_i,v_i]^{\top}$.

 For fixed principal points, \eg, $\V{c}_i = [0,0]$, $i=1,2$, there is a closed form solution for the focal lengths extracted from the given fundamenatal matrix $\V{F}$, \ie, the Bougnoux formula~\cite{bougnoux_original}. Since this formula can result in unstable or imaginary estimates, in our formulation, we allow moving the estimated principal points to avoid such instabilities. 
 For non-fixed principal points $\V{c}_i$, there are infinitely many decompositions of $\V{F}$ into the essential matrix $\V{E}$ and two calibration matrices of the form~\eqref{eq:K}. 
Thus, we formulate the problem of estimating focal lengths from $\V{F}$  as a constrained optimization, where during the optimization, the focal lengths, along with the positions of principal points in both images are optimized to minimize the following cost function:
\begin{equation}
\begin{aligned}
\min_{f_1,f_2,\V c_1,\V c_2} \quad & \sum_{i=1,2} 
w^f_i( f_i-f_{i}^{p} )^2 + w^{\V c}_i\lVert   \V c_i - \V c_i^p
 \rVert^2\\ 
\textrm{s.t.} \quad \kappa_1 = &  \quad \sigma_1(\V v_1^{\top}\omega_1^{*} \V v_1)(\V u_1^{\top}\omega_2^{*} \V u_2)+ \\
&  \quad + \sigma_2(\V v_1^{\top}\omega_1^{*} \V v_2)(\V u_2^{\top} \omega_2^{*} \V u_2) = 0 \\
\kappa_2 = &  \quad \sigma_1(\V v_1^{\top}\omega_1^{*} \V v_2)(\V u_1^{\top}\omega_2^{*} \V u_1)+ \\
&  \quad + \sigma_2(\V v_2^{\top}\omega_1^{*} \V v_2)(\V u_1^{\top} \omega_2^{*} \V u_2) = 0,   \\
\end{aligned}
\label{eq:opt}
\end{equation}
where $f_i^{p}$ and $\V c_i^p = [u_i^p,v_i^p]^{\top}$, $i=1,2$ 
are the priors for the focal lengths and the principal points, $w_i^f$ and $w_i^{\V{c}}$  are predetermined weights, and 
$\kappa_1 = 0$ and $\kappa_2 = 0$ are two Kruppa equations~\cite{lourakis_calib}\footnote{Note, that there are three Kruppa equations, but only two are linearly independent.}, which are functions of both focal lengths and principal points that appear in $\omega^{*}_i = \V K_i \V K_i^{\top}$.
The Kruppa equations ensure that for the input fundamental matrix $\V F$ and the estimated calibration matrices $\V K_1$ and $\V K_2$~\eqref{eq:K}, the matrix $\V K_1^{\top} \V F \V K_2$ is a valid essential matrix.

The cost function in~\eqref{eq:opt} has several advantages over the cost function used in~\cite{hartley_iterative}. (1) It does not contain a term with the Sampson error on correspondences, which significantly slows down the optimization in~\cite{hartley_iterative}. (2) It operates directly on focal lengths $f_i$ instead of squared focal lengths $f_i^2$. 
(3) The minimization of~\eqref{eq:opt} does not require computation or initialization using the Bougnoux formula~\cite{bougnoux_original} as used in~\cite{hartley_iterative}. Such an initialization can be very far from the ground truth focal lengths and in many scenarios can result in $f_i^2\leq 0$. Therefore, minimization of~\eqref{eq:opt} does not need a special penalty term for $f_i^2\leq 0$ as used in~\cite{hartley_iterative}.

The constrained optimization problem~\eqref{eq:opt} can be transformed into an unconstrained problem using the method of Lagrange multipliers. In this case, this leads to  the Lagrange function
$L(f_1,f_2,\V{c}_{1},\V{c}_{2},\lambda_1,\lambda_2)$:
\begin{equation}
L = \sum_{i=1,2} 
w^f_i( f_i-f_{i}^{p} )^2 + w^{\V c}_i\lVert   \V c_i - \V c_i^p
 \rVert^2 
-  2\lambda_1 \kappa_1 - 2\lambda_2 \kappa_2, 
\label{eq:Lagrangian}
\end{equation}
where $\lambda_1$ and $\lambda_2$ are the Lagrange multipliers. The constant `$-2$' is introduced for easier subsequent manipulation of the equations and it does not influence the final solution.

In this case, if $(f_1^{\star},f_2^{\star},\V{c}_{1}^{\star},\V{c}_{2}^{\star})$
 is a point of the minimum of the original constrained problem~\eqref{eq:opt}, then there exists $\lambda_1^{\star}$ and $\lambda_2^{\star}$ such that $(f_1^{\star},f_2^{\star},\V{c}_{1}^{\star},\V{c}_{2}^{\star},\lambda_1^{\star},\lambda_2^{\star})$ is a stationary point of $L$~(\ref{eq:Lagrangian}), \ie, a point where all the partial derivatives of $L$ vanish. The Lagrange function $L(f_1,f_2,\V{c}_{1},\V{c}_{2},\lambda_1,\lambda_2)$~(\ref{eq:Lagrangian}) is a function of eight unknowns. Thus, to find  all stationary points of $L$ we need to solve the following system of eight polynomial equations in eight unknowns \footnote{Note that~\eqref{eq:1-2} represents two equations for $i=1,2$ and~\eqref{eq:3-6} represents four equations since there are two vector equations for $i=1,2$.}:
 \begin{eqnarray}
\label{eq:1-2}
2 w_i^f(f_i-f_i^p) - 2 \lambda_1 \frac{\partial \kappa_1}{\partial f_i} - 2 \lambda_2 \frac{\partial \kappa_2}{\partial f_i} &= 0,  \\
\label{eq:3-6}
2 w_i^\V{c} (\V c_i-\V c_i^p) - 2 \lambda_1 (\frac{\partial \kappa_1}{\partial \V{c}_i})^{\top} - 2 \lambda_2 (\frac{\partial \kappa_2}{\partial \V c_i})^{\top} &= \V 0,  \\
\label{eq:7}
\kappa_1 &= 0, \\
\label{eq:8}
\kappa_2 &= 0.
\end{eqnarray}
Unfortunately, the system of eight equations~\eqref{eq:1-2}-\eqref{eq:8} is too complex to be efficiently solved using an algebraic method, \eg using the  Gr\"{o}bner basis method~\cite{kukelova_automatic,larsson_efficient}. 
Next, we propose an iterative method to efficiently solve this system of polynomial equations. Note that in our method, we estimate all stationary points of the Lagrange function $L$~\eqref{eq:Lagrangian}. Thus, we avoid potential problems of numerical methods that directly minimize~\eqref{eq:Lagrangian}. Such methods usually require good initialization and can get stuck in local minima.

\subsection{Iterative method}
\label{sec:method}
First, let us denote $\Delta f_i = f_i - f_i^p$, 
$\Delta \V c_i = \V c_i - \V c_i^p$, $i=1,2$.
From equations~\eqref{eq:1-2}--\eqref{eq:3-6} we have
\begin{eqnarray}
\label{eq:1-2_2}
\Delta f_i =& \frac{1}{w_i^f}(\lambda_1 \frac{\partial \kappa_1}{\partial f_i}
+ \lambda_2 \frac{\partial \kappa_2}{\partial f_i}),\\
\label{eq:3-6_2}
\Delta \V c_i =& \frac{1}{w_i^\V{c}}\left(\lambda_1 \left(\frac{\partial \kappa_1}{\partial \V{c}_i}\right)^{\top}
+ \lambda_2 \left(\frac{\partial \kappa_2}{\partial \V{c}_i}\right)^{\top}\right).
\end{eqnarray}
and the cost function in the original constrained optimization problem~\eqref{eq:opt} can be rewritten as
\begin{eqnarray}
\label{eq:minimize_delta}
e = \sum_{i=1,2} 
w^f_i \Delta f_i^2 + w^{\V c}_i\Delta \V c_i^{\top}\Delta \V c_i  \end{eqnarray}

The proposed iterative solution to equations~\eqref{eq:1-2}--\eqref{eq:8} follows the idea of iterative triangulation methods~\cite{Lindstrom-CVPR-2010, Kukelova_rad_triang}. First, let $s^{k-1} = \langle f_1^{k-1},f_2^{k-1},\V{c}_{1}^{k-1},\V{c}_{2}^{k-1}\rangle$ denote the best current estimate of focal lengths and principal points after the $(k-1)^{th}$ iteration. 
The prior values $f_{i}^{0} \equiv f_i^p$, $\V c_{i}^{0} \equiv \V c_i^p$, $i=1,2$ are used as an initialization. 
In the $k^{th}$ iteration the updated estimates $s^{k} = \langle f_1^{k},f_2^{k},\V{c}_{1}^{k},\V{c}_{2}^{k}\rangle$ are obtained by replacing the focal lengths and principal points  $\langle f_1,f_2,\V{c}_{1},\V{c}_{2}\rangle$ on the
right-hand side of equations~(\ref{eq:1-2_2})--(\ref{eq:3-6_2}) by the current best estimates $s^{k-1} = \langle f_1^{k-1},f_2^{k-1},\V{c}_{1}^{k-1},\V{c}_{2}^{k-1}\rangle$. 
In this way, we obtain the 
expressions for $\Delta f_i = \frac{1}{w_i^f}(\lambda_1 \frac{\partial \kappa_1}{\partial f_i}(s^{k-1})
+ \lambda_2 \frac{\partial \kappa_2}{\partial f_i}(s^{k-1}))$ and $\Delta \V{c}_i = \frac{1}{w_i^\V{c}}(\lambda_1 (\frac{\partial \kappa_1}{\partial \V{c}_i}(s^{k-1}))^{\top}
+ \lambda_2 (\frac{\partial \kappa_2}{\partial \V{c}_i}(s^{k-1}))^{\top})$, $i=1,2$. Here $\frac{\partial \kappa_j}{\partial f_i}(s^{k-1})$ and $\frac{\partial \kappa_j}{\partial \V c_i}(s^{k-1})$, $i,j = 1,2$ are partial derivatives of $\kappa_j$ 
evaluated at $s^{k-1} = \langle f_1^{k-1},f_2^{k-1},\V{c}_{1}^{k-1},\V{c}_{2}^{k-1}\rangle$.  
After this substitution, $\Delta f_i$ and $\Delta \V c_i$ are functions of two Lagrange multipliers $\lambda_1$ and $\lambda_2$.


Expressions $\Delta f_i$ and $\Delta \V c_i$ are, in turn, substituted into the Kruppa equations~\eqref{eq:7} and~\eqref{eq:8}. 
This is done by substituting $\Delta f_i$ and $\Delta \V{c}_i$, $i=1,2$ into 
$\omega_i^*$
using the relationship 
$f_i = \Delta f_i + f_i^p$ and $\V c_i = \Delta \V c_i + \V c_i^p$. 

The updated Kruppa equations~\eqref{eq:7} and~\eqref{eq:8}, which we denote $\kappa_1^{k}=0$ and $\kappa_2^{k}=0$, form a system of two equations of degree four in two unknowns 
$\lambda_1$ and $\lambda_2$.
This system has up to 16 real solutions and can be efficiently solved using the Gr\"{o}bner basis method~\cite{larsson_efficient}. The final solver performs an elimination of a $20 \times 36$ matrix and extracts solutions from the eigenvalues and eigenvectors of a $16 \times 16$ matrix.

From up to 16 possible solutions, a solution $\lambda_1^k$ and $\lambda_2^k$
 that minimizes $|\lambda_1| + |\lambda_2|$ is selected~\cite{Lindstrom-CVPR-2010}. 
 This solution is used to compute the $k^{th}$ iteration updates $\Delta f_i^{k}$ and $\Delta \V{c}_{i}^{k}$ using equations~\eqref{eq:1-2_2}--\eqref{eq:3-6_2} and subsequently the new estimates of focal lengths and principal points $s^{k} = \langle f_1^{k},f_2^{k},\V{c}_{1}^{k},\V{c}_{2}^{k}\rangle$. 
 
 The iterative algorithm
 stops when the relative change of error of two consecutive iterations $\frac{\left|e^{k}-e^{k-1}\right|}{e^{k}} < \epsilon$,
 for some threshold $\epsilon$.
 Note that in each iteration, the obtained solutions satisfy the Kruppa equations $\kappa_1=0$ and $\kappa_2=0$. This means that for the input fundamental matrix $\V F$ and the estimated calibration matrices $\V K_1$ and $\V K_2$~\eqref{eq:K}, the matrix $\V K_1^{\top} \V F \V K_2$ is a valid essential matrix in each iteration.

The situation with $f_1 = f_2$ can be solved using the same method. In this case, in the cost function~\eqref{eq:opt}, as well as in the Lagrangian and in~\eqref{eq:1-2}--\eqref{eq:3-6}, we only have $i = 1$.  However, since we still have two Kruppa equations $\kappa_1 = 0$ and $\kappa_2 = 0$ the final algorithm
also relies on solutions to a system of two equations of degree four in two unknowns.
The algorithm is summarized in the Supplementary material.

\def\kmo{\begingroup\setmuskip{\medmuskip}{0mu}$(k-1)$\endgroup}

\subsection{Real Focal Length Checking within RANSAC}

\label{sec:rfc}
The method proposed in~\ref{sec:method} is usually applied on a fundamental matrix $\V{F}$ that is obtained using a robust RANSAC-style estimation~\cite{ransac, magsac++, lo-ransac, degensac, vsac, prosac}.
Compared to the Bougnoux formula~\cite{bougnoux_original}, our method does not suffer from imaginary focal length estimates in the presence of noise or an error in the principal point location. Thus, our method does not need to check whether the fundamental matrix returned by the RANSAC is decomposible to real focal lengths or not. 
Nevertheless, as we observed in our experiments, fundamental matrices that result in imaginary focal lengths usually do not provide good estimates and are therefore not selected inside RANSAC as best models.

Inspired by this observation, we propose a simple modification of RANSAC that is based on rejecting models that lead to imaginary focal lengths.  Rejection is done before scoring models on all point correspondences.
This reduces the computational time required to score a model, which is unlikely to lead to the final output. 
Thus, this approach is similar to other degeneracy checking algorithms.

To reject models, we use a modified version of the Bougnoux formula~\cite{bougnoux_original} 
presented in~\cite{oleh_bp}, which has the form of a ratio of two polynomials in the elements of the fundamental matrix.
The formula returns squared focal lengths. When the sign of one of the squared focal lengths is negative, we reject the model. This approach does not require the use of SVD or other expensive matrix manipulations. This type of checking can be implemented into different variants of RANSACs and, as shown later in our experiments, it leads to reduced computational times while maintaining or even improving accuracy of pose estimates. Implementation details can be found in the Supplementary Material (SM). 



\section{Synthetic Experiments}

\begin{figure*}[t]
    \begin{subfigure}[t]{0.32\textwidth}        
    \centering
    \includegraphics[height=0.55\textwidth]{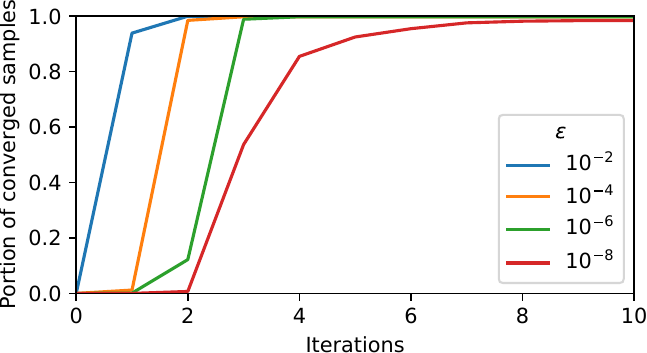}
    \caption{$\mathcal{C}(0\degree, 300)$}
    \label{fig:cvg_synth_ncp}
    \end{subfigure}\hfill%
    ~
    \begin{subfigure}[t]{0.32\textwidth}        
    \centering
    \includegraphics[height=0.55\textwidth]{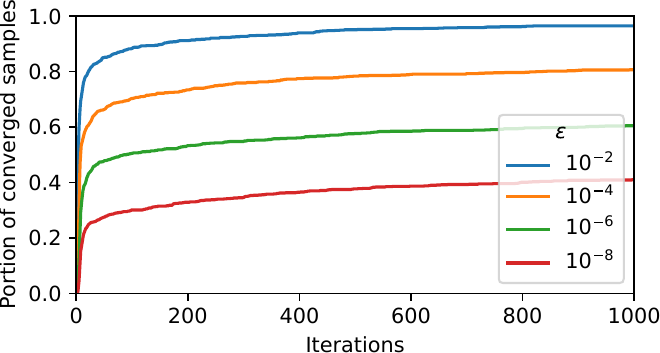}
    \caption{$\mathcal{C}(0\degree, 0)$}
    \label{fig:cvg_synth_cp}
    \end{subfigure}\hfill%
    ~
    \begin{subfigure}[t]{0.32\textwidth}        
    \centering
    \includegraphics[height=0.55\textwidth]{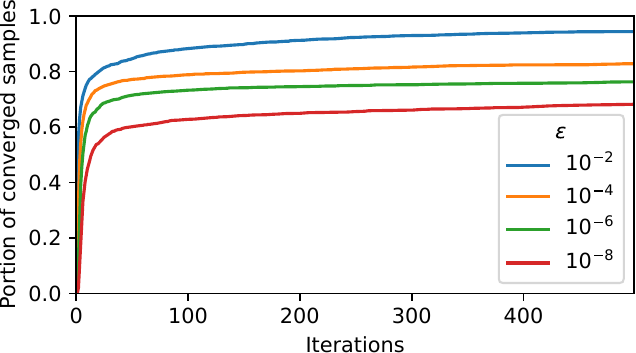}
    \caption{Phototourism~\cite{phototourism_benchmark}}
    \label{fig:cvg_real}
    \end{subfigure}
    \label{fig:convergence}
    \vspace{-2mm}
    \caption{Plots showing portion of samples for which our algorithm would converge given a threshold for the relative change of errors in successive iterations $\frac{|e^{n} - e^{n - 1}|}{e^{n}} < \epsilon$. For synthetic experiments (a) and (b) we generated 1000 samples with added noise ($\sigma_n = 1$, $\sigma_p = 10$). We set the priors as $f_1^p = 660, f_2^p = 440$. For (c) we used the Phototourism benchmark dataset~\cite{phototourism_benchmark}. See Section \ref{sec:real_eval} for details.}
\end{figure*}

We perform several synthetic experiments to evaluate various aspects of our method and compare it with the state-of-the-art in controlled configurations. Here, we only include experiments for two unknown but different focal lengths. Further experiments for the case where the focal lengths are assumed to be equal can be found in the SM.

 In all our synthetic and real experiments, we set the weights in~\eqref{eq:opt} to $w_i^f = 5 \cdot 10^{-4}$ and $w_i^{\V{c}} = 1.0$.
These weights were obtained by testing different combinations on the Brandenburg Gate validation scene from the Phototurism dataset~\cite{phototourism_benchmark} and resulted in the best performance on this scene. However, as we show also in synthetic experiments, for a wide variety of combinations of weights, we consistently obtain very good estimates and outperform the state-of-the-art methods. 
We set the weights for Hartley and Silpa-Anan's method~\cite{hartley_iterative} in the same manner to 1.0 for $w_F$, $10^{-4}$ for $w_{\V{c}}$ and $10^{-6}$ for $w_f$.

\vspace{-1.5ex}
\paragraph{Camera Setup:}
To perform the experiments, we set up a scene with two cameras. We assume $640~\times~480$ image size and principal points in image centers. The first camera has focal length $f_1 = 600$ and the second 
$f_2 = 400$. The first camera has center at the origin with the z-axis of the world coordinate system being its principal axis. We perform experiments with the second camera in different configurations denoted as $\mathcal{C}(\theta, y)$. 
For $\mathcal{C}(0\degree, y)$ the second camera center is located at coordinates $(1200, y, 600)$. The camera is rotated by $60\degree$ around its $y$-axis. 
With $\theta \neq 0\degree$ we additionally rotate the second camera around the $x$-axis by $\theta$. When $y = 0$ and $\theta = 0\degree$ the principal axes of the two cameras intersect and are thus coplanar, which is a degenerate configuration for the Bougnoux formula~\cite{bougnoux_original}.


For a given camera setup, we uniformly generate 100 random points visible by both cameras. After projecting the points into both images, we add Gaussian noise with a standard deviation $\sigma_n$ to the pixel coordinates. 
We also emulate errors in the assumed locations of the principal points. The principal points are shifted from the image centers by a distance sampled from the normal distribution with a standard deviation of $\sigma_p$.
We estimate the fundamental matrix using MAGSAC++~\cite{magsac++} as implemented in OpenCV~\cite{opencv}.

\vspace{-1.5ex}
\paragraph{Convergence:}
In the first experiment, we evaluate convergence of the proposed iterative method. Different termination criteria can be considered. We terminate the algorithm when the relative change in error in two successive iterations $\frac{|e^{n} - e^{n - 1}|}{e^{n}}$ is less than a predetermined threshold. Fig.~\ref{fig:cvg_synth_ncp} shows that for non-degenerate configuration our method converges quickly even with strict thresholds. Fig.~\ref{fig:cvg_synth_cp} shows the convergence rates of our method when the principal axes are coplanar. The degenerate configuration requires an increased number of iterations to converge with the same thresholds. However, even in a degenerate configuration, the method converges in a few iterations for most of the samples, even with strict thresholds. We also evaluate convergence on a large scale real-world dataset~\cite{phototourism_benchmark} (see section~\ref{sec:real_eval} for more details). As shown in Fig.~\ref{fig:cvg_real}, our method successfully converges for most of the samples even with a strict threshold. Using these findings, we established the maximum number of iterations for our method at 50, as further iterations demonstrated diminishing returns in terms of convergence rate.

\vspace{-1.5ex}
\paragraph{Accuracy of the Estimated Focal Lengths}
In the next experiment, we compare our method with state-of-the-art methods in terms of accuracy of the estimated focal lengths. The methods considered are the
baseline Bougnoux formula~\cite{bougnoux_original} and the iterative approaches of Hartley and Silpa-Anan~\cite{hartley_iterative} and Fetzer \etal~\cite{fetzer}. 
In the experiments, we use the same priors for Hartley and Silpa-Anan's method as for ours. We set the initial values for LM optimization for the method by Fetzer \etal to the same values as our priors.

\begin{figure*}[t]
    \centering

    \begin{tikzpicture} 
    \begin{axis}[%
    hide axis, xmin=0,xmax=0,ymin=0,ymax=0,
    legend style={draw=white!15!white, 
    line width = 1pt,
    legend  columns =-1, 
    }
    ]
    \addlegendimage{line width=4pt, Seaborn1}
    \addlegendentry{\scriptsize{Ours}};
    \addlegendimage{line width=4pt, Seaborn2}
    \addlegendentry{\scriptsize{Hartley and Silpa-Anan\cite{hartley_iterative}}};
    \addlegendimage{line width=4pt, Seaborn3}
    \addlegendentry{\scriptsize{Fetzer \etal \cite{fetzer}}};
    \addlegendimage{line width=4pt, Seaborn4}
    \addlegendentry{\scriptsize{Bougnoux\cite{bougnoux_original}}};
    \addlegendimage{white}
    \addlegendentry{};
    \addlegendimage{black,dash pattern=on 4pt off 2pt on 4pt off 2pt}
    \addlegendentry{\scriptsize{GT $f_1$}};
    \addlegendimage{black!30,dash pattern=on 4pt off 2pt on 4pt off 2pt}
    \addlegendentry{\scriptsize{Prior $f_1$}};
    \end{axis}
    \end{tikzpicture}

    \begin{subfigure}[t]{0.32\textwidth}
    \includegraphics[width=\textwidth]{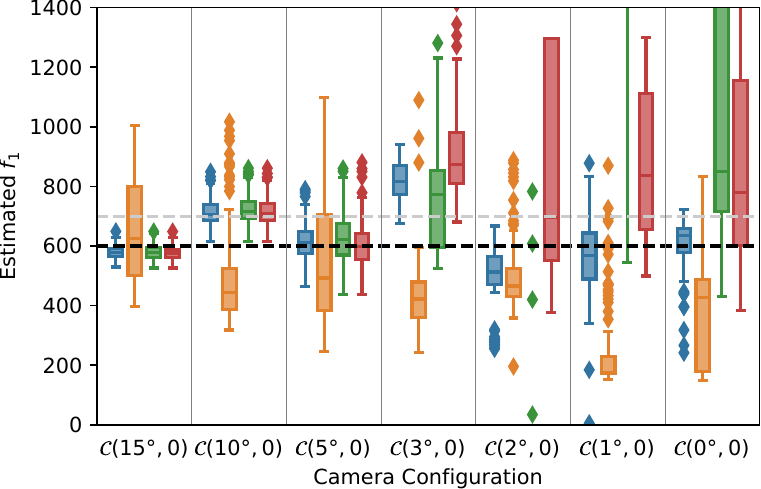}
    \vspace*{-5mm}
    \caption{}
    \label{fig:synth_uncal_coplanarity}
    \end{subfigure}\hfill%
    ~
    \centering
    \begin{subfigure}[t]{0.32\textwidth}
    \includegraphics[width=\textwidth]{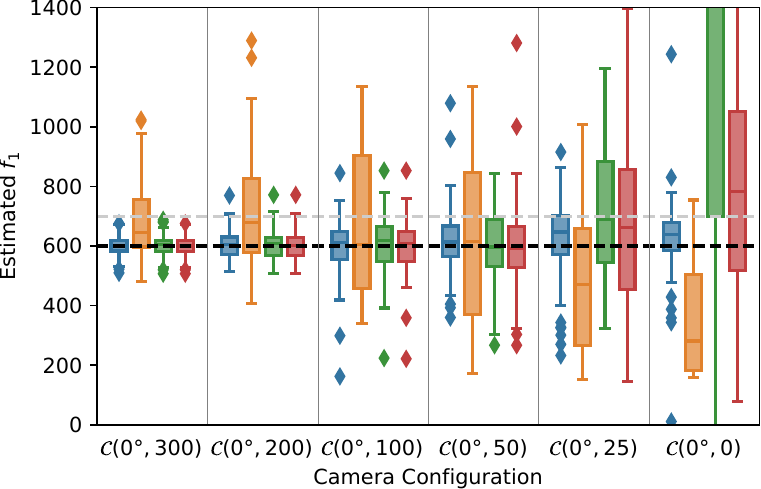}
    \vspace*{-5mm}
    \caption{}
    \label{fig:synth_uncal_coplanarity_y}
    \end{subfigure}\hfill%
    ~    
    \begin{subfigure}[t]{0.32\textwidth}
    \includegraphics[width=\textwidth]{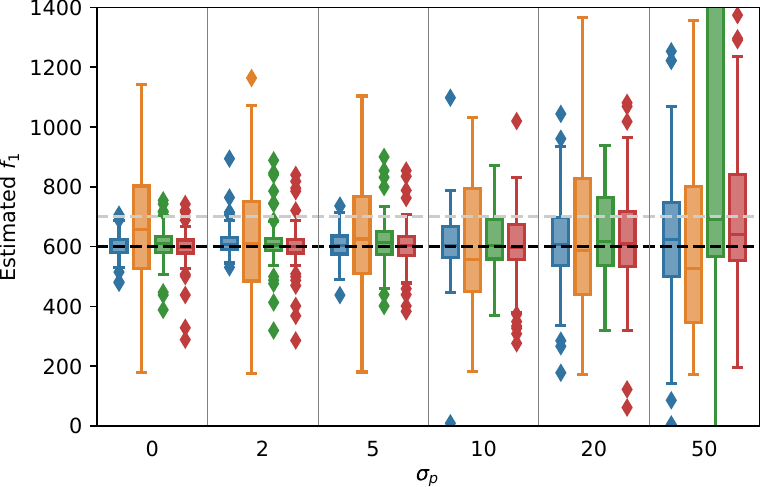}
    \vspace*{-5mm}
    \caption{}
    \label{fig:synth_uncal_principal}
    \end{subfigure}
    \hfill

    \begin{subfigure}[t]{0.32\textwidth}
    \includegraphics[width=\textwidth]{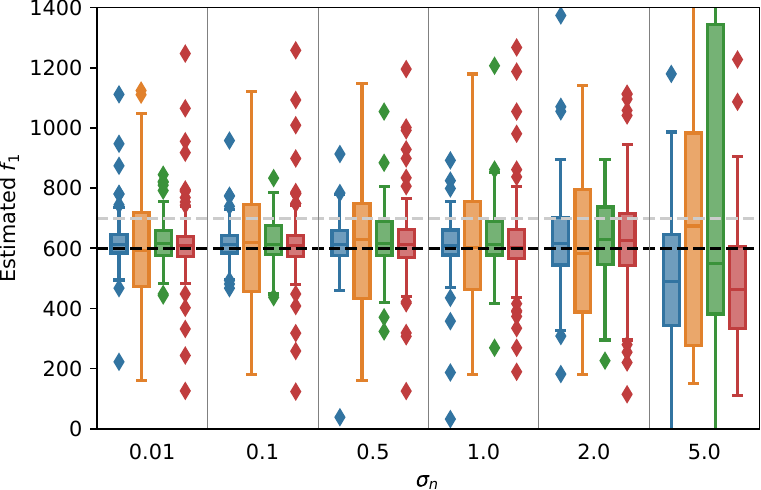}
    \vspace*{-5mm}
    \caption{}
    \label{fig:synth_uncal_noise}
    \end{subfigure}\hfill%
    ~
    \begin{subfigure}[t]{0.32\textwidth}
    \includegraphics[width=\textwidth]{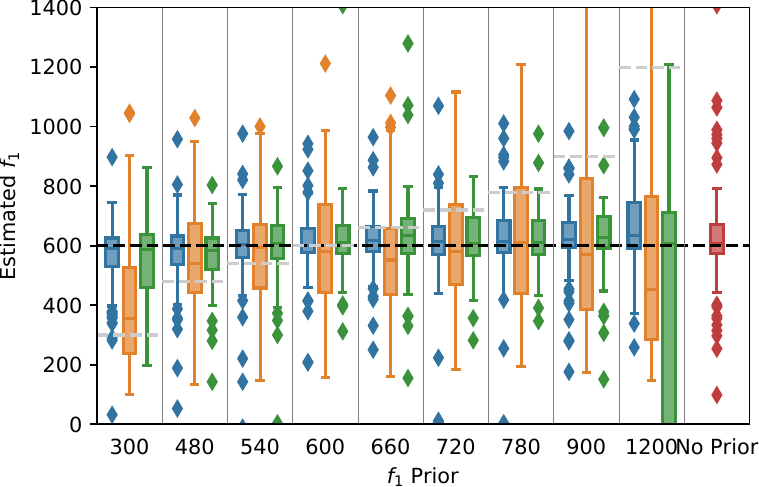}
    \vspace*{-5mm}
    \caption{}
    \label{fig:synth_uncal_priors}
    \end{subfigure}
    ~
    \begin{subfigure}[t]{0.32\textwidth}
    \includegraphics[width=\textwidth]{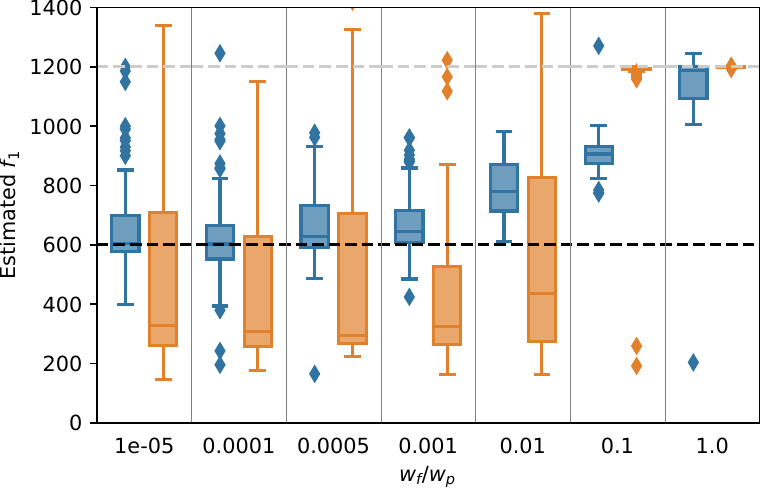}
    \vspace*{-5mm}
    \caption{}
    \label{fig:synth_uncal_weights}
    \end{subfigure}
    \hfill
    \vspace{-1ex}
    \caption{Synthetic experiments: Box plots for the estimated focal lengths of the first camera. Comparison of the methods as (a, b) the camera configuration approaches the degeneracy with coplanar principal axes, (c) we vary the standard deviation $\sigma_p$ of the error of the principal point, (d) we vary the noise added to the projected points, (e) we vary the prior for $f_1$, (f) we vary the relative weights of focal length and principal point priors. We use priors $f_1^p = 700, f_2^p = f_2 = 400$ for (a, b, c, d), $f_1^p = 1000, f_2^p = 400$ for (f),  $\sigma_n = 1$ for (a, b, c, e, f), $\sigma_p = 10$ for (a, b, d, e, f). For (c, d, e, f) we randomly sample the configuration $\mathcal{C}(\theta, y)$ with $\theta \in \left[-15\degree, 15 \degree \right]$ and $y \in \left[-200, 200\right]$.}
    \label{fig:synth_coplanarity}   
\end{figure*}

Fig.~\ref{fig:synth_uncal_coplanarity} and \ref{fig:synth_uncal_coplanarity_y} show how the transition to a degenerate configuration with coplanar principal axes affects the accuracy of the evaluated methods. Our method performs well even close to and in degenerate configurations. In comparison, the other evaluated methods fail to provide consistent results when the camera position approaches the degenerate configuration. 

A similar trend is visible in Fig.~\ref{fig:synth_uncal_principal} where we show that an increase in the difference between the assumed and ground truth position of the principal point results in a worsening performance of the Bougnoux formula and the method by Fetzer~\etal. Our method is capable of compensating for the error in the principal point as the principal point is not assumed to be known exactly. The performance of Hartley and Silpa-Anan's method stays consistent for similar reasons, although it performs worse than ours.

In Fig.~\ref{fig:synth_uncal_noise}, we show the performance of the methods under varying levels of added image noise. Our method demonstrates a higher robustness to noise, resulting in reduced inter-sample deviation and a more consistent median estimate across the various noise levels. Fig.~\ref{fig:synth_uncal_priors} shows how the different methods respond to the priors for focal length, or in the case of the method by Fetzer~\etal to an initialization for the iterative process. Fig.~\ref{fig:synth_uncal_weights} shows how the setting of weights for our and Hartley and Silpa-Anan's method affects the accuracy of the estimated focal lengths.

\section{Real-world Experiments}

\begin{figure}
    \centering
    \includegraphics[width=0.95\linewidth]{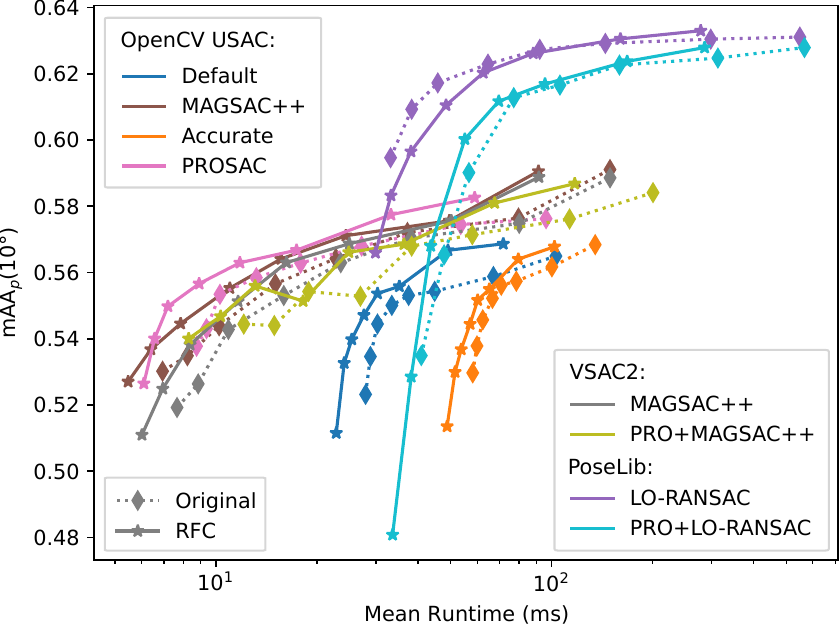}
    \vspace{-2ex}
    \caption{The plot shows mAA$_p$ achieved on the Phototourism datset~\cite{phototourism_benchmark} for different RANSAC implementations \cite{opencv, vsac, poselib} with and without real focal length checking (RFC) under varying number of iterations. To calculate the poses we use the ground truth focal lengths. Further details are provided in the SM.}
    \label{fig:rfc}
\end{figure}

\begin{figure*}

    \center
    \begin{tikzpicture} 
    \begin{axis}[%
    hide axis, xmin=0,xmax=0,ymin=0,ymax=0,
    legend style={draw=white!15!white, 
    line width = 1pt,
    legend  columns =-1, 
    }
    ]
    \addlegendimage{Seaborn1}
    \addlegendentry{\scriptsize{Ours}};
    \addlegendimage{Seaborn2}
    \addlegendentry{\scriptsize{Hartley and Silpa-Anan\cite{hartley_iterative}}};
    \addlegendimage{Seaborn3}
    \addlegendentry{\scriptsize{Fetzer \etal \cite{fetzer}}};
    \addlegendimage{Seaborn4}
    \addlegendentry{\scriptsize{Bougnoux\cite{bougnoux_original}}};
    
    \addlegendimage{Seaborn7}
    \addlegendentry{\scriptsize{Sturm \cite{sturm_single}}};
    \addlegendimage{Seaborn6}
    \addlegendentry{\scriptsize{Minimal 6-point \cite{stewenius_single_6pt}}};
    
    \addlegendimage{Seaborn5}
    \addlegendentry{\scriptsize{Prior}};
    
    \addlegendimage{white}
    \addlegendentry{};
    \addlegendimage{black!30}
    \addlegendentry{\scriptsize{RFC}};
    \addlegendimage{black!30,dash pattern=on 4pt off 2pt on 4pt off 2pt}
    \addlegendentry{\scriptsize{Original}};
    \end{axis}
    \end{tikzpicture}    
    
    \begin{subfigure}[t]{0.32\textwidth}
    \includegraphics[width=\textwidth]{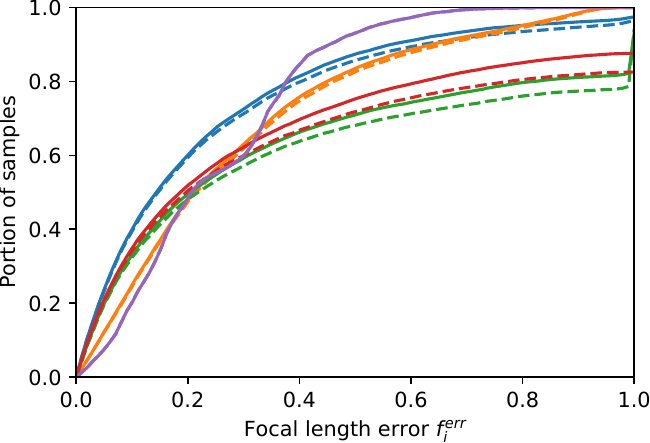}
    \caption{Phototourism~\cite{phototourism_benchmark}}
    \label{fig:real_eval_pt}
    \end{subfigure}\hfill%
    ~
    \begin{subfigure}[t]{0.32\textwidth}
    \includegraphics[width=\textwidth]{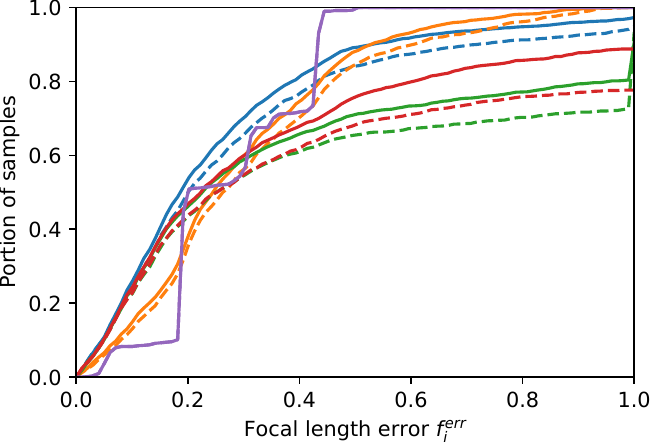}
    \caption{Aachen Day-Night v1.1~\cite{aachen_dn_extended}}
    \label{fig:real_eval_aachen}
    \end{subfigure}\hfill%
    ~
    \begin{subfigure}[t]{0.32\textwidth} 
    \includegraphics[width=\textwidth]{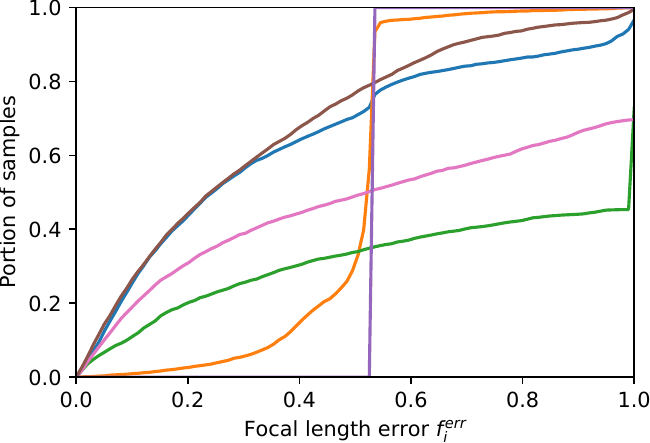}
    \caption{ETH3D Multiview~\cite{eth3d_multiview}}
    \label{fig:real_eval_single}
    \end{subfigure}

    \vspace{-1ex}    
    \caption{Plots showing  the portion of samples for which the estimated focal lengths were below a given $f_i^{err}$ threshold. For (a) and (b) we assume different focal lengths for the two cameras, for (c) the focal lengths are assumed to be equal.}
    \label{fig:real_eval_f}
    
\end{figure*}

\label{sec:real_eval}

\begin{table*}[]
    \centering
    \resizebox{\textwidth}{!}{%
    \begin{tabular}{|*{14}{c|}}   
\cline{3-14}
\nocell{2}  & \multicolumn{6}{|c|}{Phototourism~\cite{phototourism_benchmark}} & \multicolumn{6}{|c|}{Aachen Day-Night v1.1~\cite{aachen_dn_extended}} \\ \hline
\multirow{2}{*}{Method} & \multirow{2}{*}{RFC} &  Median   & \multicolumn{2}{|c|}{mAA$_p$} &   Median     &  \multicolumn{2}{|c|}{mAA$_f$} &   Median   & \multicolumn{2}{|c|}{mAA$_p$} &   Median     &  \multicolumn{2}{|c|}{mAA$_f$}\\ \cline{4-5} \cline{7-8}\cline{10-11}\cline{13-14}
                       &                       &  $p_{err}$& 10\degree  & 20\degree        &   $f_{err}$  & 0.1                 & 0.2      &   $p_{err}$& 10\degree  & 20\degree        &   $f_{err}$  & 0.1                 & 0.2     \\ 
\hline
\multirow{2}{*}{Ours} 	& 	                    & 	 6.43\degree & 	 40.48 &	 56.01 & 	 	 0.146 &	 24.15 &	 37.72 &  	 9.52\degree & 	 27.62 &	 46.28 & 	 	 0.199 &	 12.61 &	 25.59 \\  
                       & \checkmark  	        & 	 	\textbf{6.29\degree} & 	 \textbf{41.03} &	 \textbf{56.78} & 	 	 \textbf{0.143} &	 \textbf{24.47} &	 \textbf{38.19} &  	 	\textbf{8.78\degree} & 	 \textbf{29.29} &	 \textbf{48.13} & 	 	 \textbf{0.188} &	 \textbf{13.20} &	 \textbf{27.33} \\ \hline
\multirow{2}{*}{Hartley \cite{hartley_iterative}}& & 	9.19\degree & 	 30.15 &	 46.94 & 	 	 0.215 &	 13.35 &	 25.45 &  	 12.19\degree & 	 21.10 &	 38.74 & 	 	 0.263 &	 6.77 &	 14.94 \\  
  	              & \checkmark  	                  & 	 9.00\degree & 	 30.61 &	 47.69 & 	 	 0.211 &	 13.49 &	 25.76 &  	 11.37\degree & 	 22.12 &	 40.61 & 	 	 0.245 &	 7.66 &	 16.55 \\ \hline
\multirow{2}{*}{Fetzer \cite{fetzer}} 	 & 	                               & 	9.40\degree & 	 33.00 &	 47.25 & 	 	 0.218 &	 19.97 &	 30.84 &  	 12.33\degree & 	 24.34 &	 39.69 & 	 	 0.251 &	 12.07 &	 23.62 \\  
   	                & \checkmark  	                   & 	 8.94\degree & 	 33.64 &	 48.51 & 	 	 0.204 &	 20.38 &	 31.62 &  	 10.66\degree & 	 25.62 &	 42.16 & 	 	 0.226 &	 12.16 &	 24.84 \\ \hline
\multirow{2}{*}{Bougnoux \cite{bougnoux_original}}  	 & 	                             & 	 7.55\degree & 	 37.17 &	 52.70 & 	 	 0.197 &	 21.09 &	 32.44 &  	 10.25\degree & 	 26.73 &	 45.07 & 	 	 0.250 &	 11.96 &	 23.72 \\  
                    	 & \checkmark  	                 & 	 7.39\degree & 	 37.57 &	 53.21 & 	 	 0.187 &	 21.51 &	 33.25 &  	 9.69\degree & 	 27.25 &	 45.61 & 	 	 0.223 &	 12.44 &	 25.17 \\ \hline 
\multirow{2}{*}{Prior} 	 & 	                                & 	 11.17\degree & 	 22.63 &	 41.98 & 	 	 0.206 &	 9.52 &	 22.75 &  	 13.00\degree & 	 17.06 &	 37.27 & 	 	 0.195 &	 4.43 &	 10.02 \\  
	 & \checkmark  	                    & 	 11.05\degree & 	 22.73 &	 42.31 & 	 	 0.206 &	 9.52 &	 22.75 &  	 12.73\degree & 	 17.68 &	 38.05 & 	 	 0.195 &	 4.43 &	 10.02 \\ \hline
\multirow{2}{*}{GT intrinsics} 	 & 	                                   & 	 2.85\degree & 	 59.78 &	 70.80 & 	 	\multirow{2}{*}{---} &	 \multirow{2}{*}{---} &	 \multirow{2}{*}{---} &  	 5.45\degree & 	 45.99 &	 58.66 & 	 	 \multirow{2}{*}{---} &	 \multirow{2}{*}{---} &	 \multirow{2}{*}{---} \\  
         	 & \checkmark  	                       & 	 2.82\degree & 	 60.05 &	 71.18 & 	 	  &	  &	  &  	 4.25\degree & 	 51.76 &	 64.82 & 	 	  &	  &	  \\ \hline
    \end{tabular}}
    \vspace{-1ex}
    \caption{Median errors for poses ($p_{err}$) and focal lengths ($f_i^{err}$) and mean average accuracy scores for poses (mAA$_p$) and estimated focal lengths (mAA$_f$) on 12 scenes from the Phototourism dataset~\cite{phototourism_benchmark} and the Aachen Day-Night v1.1 dataset~\cite{aachen_dn_extended}. RFC denotes real focal length checking as described in subsection \ref{sec:rfc}.}
    \label{tab:real_eval}
\end{table*}

\vspace{-1.5ex}
\paragraph{Datasets:}
We assess the real-world performance of our method using two extensive datasets. The first dataset, referred to as the \textbf{Phototourism dataset}, was originally introduced in~\cite{phototourism_benchmark} to serve as a robust benchmark to evaluate structure-from-motion pipelines. The dataset contains 25 scenes of landmarks. Large sets of crowd-sourced images are available for each scene. For 13\footnote{Originally, reconstructions were available for 16 of the scenes, but 3 of the scenes were later dropped from the dataset due to data inconsistencies.} of the scenes a COLMAP~\cite{colmap} reconstruction is provided to serve as ground truth for camera intrinsics and extrinsics. To evaluate the estimated focal lengths, we randomly sample 1000 pairs of images for each scene. We only consider pairs with sufficient co-visibility as defined in~\cite{phototourism_benchmark}. We used the Brandenburg Gate scene to serve as a validation set to determine the optimal setting of the weights $w_i^f$ and $w_i^{\V c}$ in~\eqref{eq:opt}.

The second dataset, known as the \textbf{Aachen Day-Night v1.1 Dataset}, is an extension of the previously established Aachen Day-Night dataset~\cite{aachen_dn, aachen_original}. This dataset incorporates a reference COLMAP~\cite{colmap} model that has been reconstructed from a comprehensive collection of 6,697 images captured by a variety of cameras and smartphones. For evaluation, we randomly sample 1000 pairs of images that adhere to the same co-visibility criteria as those defined for the Phototourism dataset.

\vspace{-1.5ex}
\paragraph{Fundamental Matrix Estimation:}
We use LoFTR~\cite{loftr} to obtain point correspondences. We perform inference on images resized so that the largest dimension is equal to 1024 px. 
To estimate the fundamental matrix, we use MAGSAC++~\cite{magsac++} 
with the epipolar threshold set to 3 px. 

We also perform experiments with and without utilizing real focal length checking (RFC) as described in subsection~\ref{sec:rfc}. 
Fig~\ref{fig:rfc} shows that on the Phototourism dataset~\cite{phototourism_benchmark} the fundamental matrices obtained with RFC generally lead to more accurate pose estimates while also significantly reducing computational times across various RANSAC implementations. 
Implementation details and further experiments are provided in SM.


\vspace{-1.5ex}
\paragraph{Compared Methods:}
%
We compare our method with three state-of-the-art approaches. We evaluate the Bougnoux formula~\cite{bougnoux_original}, Hartley and Silpa-Anan's iterative method~\cite{hartley_iterative} and the iterative method by Fetzer~\etal~\cite{fetzer}. 

For all images, we consider the principal point to lie in their center which is a commonly used assumption. With the exception of the Bougnoux formula, all methods require priors, or initial estimates, for the focal lengths. Following common practice, as established in COLMAP~\cite{colmap}, we initialize the priors for focal lengths at $1.2$ times the maximum dimension of the image (i.e., either width or height). This initial setting implies an approximate 50-degree field of view for the cameras.


\paragraph{Metrics:}
To evaluate the accuracy of the output focal lengths, we define relative focal length error as
\begin{equation}
    f_i^{err} = \frac{|f_i^{est} - f_i^{gt}|}{\text{max}\left(f_i^{est}, f_i^{gt} \right)},
\end{equation}
with $f_i^{est}$ and $f_i^{gt}$ denoting the estimated and ground truth focal lengths respectively, with $i \in \left\{1, 2\right\}$ denoting the first or the second camera. To gauge the accuracy of the poses, we employ the Mean Average Accuracy (mAA) metric, as originally proposed in~\cite{phototourism_benchmark}. The basis for this metric is the pose error $(p_{err})$ which is calculated as the maximum of the rotation and translation error in degrees.

Furthermore, we employ the mAA metric to evaluate and compare the accuracy of the estimated focal lengths, based on the $f_i^{err}$ curves as presented in Figure \ref{fig:real_eval_f}. To distinguish between the two applications of the mAA metric, we denote them as mAA$_p$ for pose and mAA$_f$ for focal lengths.

\vspace{-2.0ex}
\paragraph{Results:}
The median errors and mean average accuracy scores for both pose and focal lengths are reported in Table~\ref{tab:real_eval}. On both the Phototourism and the Aachen dataset, our method significantly outperforms others in terms of the accuracy of the estimated focal lengths. The superiority of our method in terms of focal length accuracy can also be seen in Fig.~\ref{fig:real_eval_pt} and \ref{fig:real_eval_aachen}. In terms of pose accuracy, our method also provides more accurate results than all compared methods.

The results also show that performing RFC for real focal lengths within RANSAC as proposed in subsection~\ref{sec:rfc} generally leads to improvements in terms of both the pose and focal length accuracy across the compared methods.

\vspace{-2.0ex}
\paragraph{Computational Speed:}

\begin{table}
\resizebox{\linewidth}{!}{
\begin{tabular}{|c|c|c|c|c|}
\hline
Method & Library & RFC & Mean Time (ms) & Mean iterations \\ \hline
\multirow{4}{*}{Ours} 	 &  \multirow{2}{*}{Matlab} &          & 	 22.42 & 	 24.33 \\ \cline{4-5}
	      &     & \checkmark  	                 & 	 14.95 & 	 22.53 \\ \cline{2-5}
            &	\multirow{2}{*}{Eigen (C++)}  & 	                         & 	 3.59 & 	 23.63 \\ \cline{4-5}
            &	  & \checkmark  	             & 	 3.44 & 	 22.49 \\ \hline
\multirow{2}{*}{Hartley~\cite{hartley_iterative}} & \multirow{2}{*}{PyTorch~\cite{pytorch}}	& 	                              & 	 659.61 & 	 85.38 \\ \cline{4-5}
 	&  & \checkmark  	                  & 	 663.98 & 	 86.14 \\ \hline
\multirow{2}{*}{Fetzer~\cite{fetzer}} &  \multirow{2}{*}{PyTorch~\cite{pytorch}}	 & 	                               & 	 197.89 & 	 0.00 \\ \cline{4-5}
 &	 & \checkmark  	                   & 	 191.75 & 	 40.56 \\ \hline
\multirow{2}{*}{Bougnoux~\cite{bougnoux_original}}  &\multirow{2}{*}{NumPy~\cite{numpy}}	 & 	                             & 	 0.27 & 	 --- \\ \cline{4-5}
 	 & & \checkmark  	                 & 	 0.25 & 	 --- \\ \hline
\end{tabular}}
\vspace{-1ex}
\caption{Mean computation times and number of iterations for the compared methods on the Phototourism dataset~\cite{phototourism_benchmark}. Measurements were performed on Intel i7-11800H. The times shown do not include the estimation of the fundamental matrices.}
\label{tab:speed}
\end{table}

Table~\ref{tab:speed} shows the mean computational times for the various methods. Our method is significantly faster than other iterative methods. Note that the implementation of our method is not optimized and there is still some space for speed-up.

\vspace{-2.0ex}
\paragraph{Equal Focal Lengths:}
\begin{table}
\resizebox{\linewidth}{!}{
\begin{tabular}{|c|c|c|c|c|c|c|c|}
\hline
\multirow{2}{*}{Method} &  Median   & \multicolumn{2}{|c|}{mAA$_p$} &   Median     &  \multicolumn{2}{|c|}{mAA$_f$} \\ \cline{3-4} \cline{6-7}
                        &  $p_{err}$& 10\degree  & 20\degree        &   $f_{err}$  & 0.1                 & 0.2 \\ \hline
Ours                                        &     \textbf{16.50\degree} & 	 16.70 &	 32.24 & 	 	 0.244 &	 13.69 &	 25.01 \\ \hline
Hartley \cite{hartley_iterative}            &     21.51\degree & 	 12.38 &	 24.26 & 	 	 0.523 &	 0.46 &	 1.12 \\ \hline
Fetzer \cite{fetzer}                        &     70.40\degree & 	 2.31 &	 5.20 & 	 	 0.999 &	 6.86 &	 11.64 \\ \hline
Sturm \cite{sturm_single}                   &     17.98\degree & 	 \textbf{20.64} &	 \textbf{33.30} & 	 	 0.524 &	 10.78 &	 18.46 \\ \hline
Minimal \cite{stewenius_single_6pt}         &     36.18\degree & 	 16.42 &	 25.74 & 	 	 \textbf{0.23}9 &	 \textbf{15.04} &	 \textbf{25.89 } \\ \hline
Prior                                       &     24.59\degree & 	 11.24 &	 20.92 & 	 	 0.528 &	 0.00 &	 0.00 \\ \hline
GT intrinsics                               &     6.39\degree & 	 41.73 &	 54.16 &           --- &     --- &     --- \\ \hline
\end{tabular}}
\vspace{-1ex}
\caption{Mean average accuracy scores for poses (mAA$_p$) and estimated focal lengths (mAA$_f$) on 12 scenes from the ETH3D High Resolution Multiview dataset~\cite{eth3d_multiview}. The focal lengths of both cameras are assumed to be equal. The last row includes reference results for pose accuracy when ground truth intrinsics are used.}
\label{tab:single_eval}
\end{table}

We also evaluate our method for the case when both focal lengths can be assumed to be equal. In this case, we use the modified version of our iterative algorithm from Section~\ref{sec:method}. 
To perform the evaluation, we use the ETH3D Multiview dataset~\cite{eth3d_multiview}. We use all image pairs satisfying the co-visibility criterion for each of the 12 scenes from the test set, yielding 3830 total pairs. For comparison, we modified the method by Fetzer~\etal to optimize for a single focal length. We use Hartley and Silpa-Anan's method as usual, but on output we average the two focal lengths produced. Instead of the Bougnoux formula, we use the formula proposed by Sturm~\cite{sturm_single} which is specific to the case of equal focal lengths. We also evaluate the minimal 6-point algorithm~\cite{stewenius_single_6pt} implemented within LO-RANSAC~\cite{lo-ransac, poselib}. Further details and additional experiments are available in SM.

The results are shown in Table~\ref{tab:single_eval} and Fig.~\ref{fig:real_eval_single}. Our method provides superior results to other methods based on the decomposition of the fundamental matrix, and performs comparably to the approach based on the minimal solver~\cite{stewenius_single_6pt} in terms of estimated focal lengths and shows minor improvements in terms of pose accuracy. 
Our method thus provides a good alternative to the 6-point minimal solver in the case when the focal lengths of the two cameras are known to be equal.

\section{Conclusion}

We address the important problem of robust self-calibration of the focal lengths of two cameras from a given fundamental matrix. Our new efficient iterative method shows substantial improvements over existing techniques. 
Synthetic experiments show that our method performs reliably even when the cameras are in or close to degenerate configurations. In real-world evaluations on two large-scale datasets, our approach demonstrates superior accuracy in terms of estimated poses and focal lengths, even when we use inaccurate initial priors, while also being faster than competing iterative approaches.

We have additionally proposed to perform a computationally simple check within the standard 7-point algorithm to remove models that lead to imaginary focal lengths. Real-world experiments show that this approach not only decreases computational time of the whole RANSAC pipeline, but, in general, also leads to improved pose and focal length accuracy across multiple RANASC variants and implementations. 

{\small
\bibliographystyle{ieee_fullname}
\bibliography{egbib}
}

\end{document}